\documentclass[a4paper,11pt]{article}
\usepackage{times}
\usepackage{url}
\usepackage{latexsym}
\usepackage{bigfoot}
\usepackage{comment}
\usepackage{graphicx}
\usepackage{caption}
\usepackage{subcaption}
\usepackage{amsmath}
\usepackage{color}
\usepackage{wrapfig}
\usepackage{booktabs}       

\usepackage{eacl2017}

\eaclfinalcopy 
\def\eaclpaperid{98} 

\title{Very Deep Convolutional Networks \\ for Text Classification}

\author{
Alexis Conneau \\
Facebook AI Research \\
\texttt{aconneau@fb.com} \\
\And
Holger Schwenk \\
Facebook AI Research \\
\texttt{schwenk@fb.com} \\
\And
Yann Le Cun \\
Facebook AI Research \\
\texttt{yann@fb.com} \\
\AND
Lo\"{\i}c Barrault \\
LIUM, University of Le Mans, France \\
\texttt{loic.barrault@univ-lemans.fr} \\
}

\newcommand{\eg}{e.g. }
\date{}

\usepackage{eacl2017}
\begin{document}
\maketitle
\begin{abstract}
The dominant approach for many NLP tasks are recurrent neural networks, in
particular LSTMs, and convolutional neural networks. However, these
architectures are rather shallow in comparison to the deep convolutional
networks which have pushed the state-of-the-art in computer vision.  We present a new
architecture (VDCNN) for text processing which operates directly at the character level
and uses only small convolutions and pooling operations.
We are able to show that the performance of this model increases with the
depth: using up to 29 convolutional layers, we report improvements
over the state-of-the-art on several public text classification tasks.  To the
best of our knowledge, this is the first time that very deep convolutional nets
have been applied to text processing.
\end{abstract}

\section{Introduction}
\begin{table*}[!ht]
  \centering
  \begin{tabular}{|lp{1.4cm}p{12cm}|}
    \hline
    Dataset & Label & Sample  \\
    \hline
    Yelp P. & +1 & ”Been going to Dr. Goldberg for over 10 years. I think I was one of his 1st patients when he started at MHMG. He’s been great over the years and is really all about the big picture. [...]” \\
    \hline
    Amz P. & 3(/5) & I love this show, however, there are 14 episodes in the first season and this DVD only shows the first eight. [...]. I hope the BBC will release another DVD that contains all the episodes, but for now this one is still somewhat enjoyable.\\
    \hline
    Sogou & "Sports" & ju4 xi1n hua2 she4  5 yue4 3 ri4 , be3i ji1ng 2008 a4o yu4n hui4 huo3 ju4 jie1 li4 ji1ng guo4 shi4 jie4 wu3 da4 zho1u 21 ge4 che2ng shi4 \\ \hline
    Yah. A. & "Computer, Internet" & "What should I look for when buying a laptop? What is the best brand and what's reliable?","Weight and dimensions are important if you're planning to travel with the laptop. Get something with at least 512 mb of RAM. [..] is a good brand, and has an easy to use site where you can build a custom laptop."\\\hline
  \end{tabular}
    \caption{Examples of text samples and their labels.}
      \label{figresults}
\end{table*}
The goal of natural language processing (NLP) is to process text with
computers in order to analyze it, to extract information and eventually to
represent the same information differently.  We may want to associate
categories to parts of the text (\eg POS tagging or sentiment analysis),
structure text differently (\eg  parsing), or convert it to some other form
which preserves all or part of the content (\eg machine translation,
summarization).  The level of granularity of this processing can range from
individual characters to subword units \cite{sennrich2015neural} or words up to whole sentences or even paragraphs.

After a couple of pioneer works
(\newcite{Yoshua:2001:lm_nips}, \newcite{collobert:2008:icml_nlp}, \newcite{collobert:2011:jmlt_nlp}
among others), the use of neural networks for NLP applications is 
attracting huge interest in the research community and they are 
systematically applied to all NLP tasks.  However, while the use of (deep)
neural networks in NLP has shown very good results for many tasks, it seems that
they have not yet reached the level to outperform the state-of-the-art by a
large margin, as it was observed in computer vision and speech recognition.

Convolutional neural networks, in short \textit{ConvNets}, are very successful
in computer vision. In early approaches to computer vision, handcrafted
features were used, for instance \textit{``scale-invariant feature transform (SIFT)''}\cite{lowe2004distinctive},
followed by some classifier.  The fundamental idea of ConvNets\cite{lecun1998gradient} is to consider
feature extraction and classification as one jointly trained task.  This idea
has been improved over the years, in particular by using many layers of
convolutions and pooling to sequentially extract a \textit{hierarchical representation}\cite{zeiler2014visualizing} of the input.  The best networks are using more than 150 layers as in
\cite{He:2015:resnet,He:2016:preresnet}.

Many NLP approaches consider words as basic units. An important step was the
introduction of continuous representations of words\cite{bengio2003neural}. These \textit{word
embeddings} are now the state-of-the-art in NLP.
However, it is less clear how we should best represent a sequence of words, \eg
a whole sentence, which has complicated syntactic and semantic relations. In
general, in the same sentence, we may be faced with local and long-range
dependencies.  Currently, the main-stream approach is to consider a sentence as
a sequence of tokens (characters or words) and to process them with a recurrent
neural network (RNN).  Tokens are usually processed in sequential order, from
left to right, and the RNN is expected to \textit{``memorize''} the whole
sequence in its internal states.  The most popular and successful RNN variant
are certainly LSTMs\cite{hochreiter1997long}~--~there are many works
which have shown the ability of LSTMs to model long-range dependencies in NLP
applications, \eg  \cite{Sundermeyer:2012:is_lstm,Sutskever:2014:nips_nntrans}
to name just a few.  However, we argue that LSTMs are generic learning machines
for sequence processing which are lacking task-specific structure.

We propose the following analogy. 
It is well known that a fully connected one
hidden layer neural network can in principle learn any real-valued function,
but much better results can be obtained with a deep problem-specific
architecture which develops hierarchical representations. By these means, the
search space is heavily constrained and efficient solutions can be learned with
gradient descent.  ConvNets are namely adapted for computer vision because of the compositional structure of an image. Texts have similar properties : characters combine to form n-grams, stems, words, phrase, sentences etc.

%

We believe that a challenge in NLP is to develop deep architectures which
are able to learn hierarchical representations of whole sentences, jointly with
the task.  In this paper, we propose to use deep architectures of many
convolutional layers to approach this goal, using up to 29 layers.  The design of our
architecture is inspired by recent progress in computer vision, in particular
\cite{msr:2016:iclr:vgg,He:2015:resnet}.

This paper is structured as follows. There have been previous attempts to
use ConvNets for text processing. We summarize the previous works in the next
section and discuss the relations and differences.  Our architecture is
described in detail in section~\ref{SectArch}.  We have evaluated our approach
on several sentence classification tasks, initially proposed by
\cite{Zhang:2015_nips:text_convnet}. These tasks and our experimental results
are detailed in section~\ref{SectExp}.  The proposed deep convolutional network
shows significantly better results than previous ConvNets approach.  The paper concludes with a
discussion of future research directions for very deep approach in NLP.

\section{Related work}
\label{SectRelated}

There is a large body of research on sentiment analysis, or more generally on
sentence classification tasks.  Initial approaches followed the classical two
stage scheme of extraction of (handcrafted) features, followed by a
classification stage. Typical features include bag-of-words or $n$-grams, and
their TF-IDF.  These techniques have been compared with ConvNets by
\cite{Zhang:2015_nips:text_convnet,zhang2015text}.  We use the same corpora for our
experiments.
More recently, words or characters, have been projected into a low-dimensional
space, and these embeddings are combined to obtain a fixed size representation
of the input sentence, which then serves as input for the classifier.  The
simplest combination is the element-wise mean. This usually performs badly
since all notion of token order is disregarded.

Another class of approaches are recursive neural networks.
The main idea is to use an external tool,
namely a parser, which specifies the order in which the word embeddings are
combined.  At each node, the left and right context are combined using weights
which are shared for all nodes~\cite{Socher:2011_emnlp:recur_autoenc}.  The
state of the top node is fed to the classifier.  A recurrent neural network
(RNN) could be considered as a special case of a recursive NN: the combination
is performed sequentially, usually from left to right.  The last state of the
RNN is used as fixed-sized representation of the sentence, or eventually a
combination of all the hidden states.

First works using convolutional neural networks for NLP appeared in
\cite{collobert:2008:icml_nlp,collobert:2011:jmlt_nlp}.  They have been
subsequently applied to sentence
classification~\cite{Kim:2014_emnlp:text_convnet,Kalchbrenner:2014:convNNMT,Zhang:2015_nips:text_convnet}.
We will discuss these techniques in more detail below.  If not otherwise
stated, all approaches operate on words which are projected into a
high-dimensional space.

A rather shallow neural net was proposed
in~\cite{Kim:2014_emnlp:text_convnet}: one convolutional layer (using
multiple widths and filters) followed by a max pooling layer over time. The
final classifier uses one fully connected layer with drop-out.  Results are
reported on six data sets, in particular Stanford Sentiment Treebank (SST).
A similar system was proposed in ~\cite{Kalchbrenner:2014:convNNMT}, but using
five convolutional layers. An important difference is also the introduction of
multiple \textit{temporal k-max pooling} layers. This allows to detect the $k$ most important
features in a sentence, independent of their specific position, preserving
their relative order. The value of $k$ depends on the length of the sentence
and the position of this layer in the network.
\cite{Zhang:2015_nips:text_convnet} were the first to perform sentiment
analysis entirely at the character level. Their systems use up to six
convolutional layers, followed by three fully connected classification layers.
Convolutional kernels of size 3 and 7 are used, as well as simple max-pooling
layers.  Another interesting aspect of this paper is the introduction of
several large-scale data sets for text classification. We use 
the same experimental setting (see section~\ref{SectData}).
The use of character level information was also proposed by
\cite{Santos:2014_coling:text_convnet}: all the character
embeddings of one word are combined by a max operation and they are then jointly
used with the word embedding information in a shallow architecture. In parallel to our work, \cite{yang2016hierarchical} proposed a based hierarchical attention network for document classification that perform an attention first on the sentences in the document, and on the words in the sentence. Their architecture performs very well on datasets whose samples contain multiple sentences.

In the computer vision community, the combination of recurrent and
convolutional networks in one architecture has also been investigated, with the
goal to \textit{``get the best of both worlds''},
\eg~\cite{Pinhero:2014_icml:conv+rnn}.
The same idea was recently applied to
sentence classification~\cite{Xiao:2016_arxiv:conv+rnn}. A convolutional
network with up to five layers is used to learn high-level features which serve
as input for an LSTM. The initial motivation of the authors was to obtain the
same performance as~\cite{Zhang:2015_nips:text_convnet} with networks which
have significantly fewer parameters. They report results very close to those
of~\cite{Zhang:2015_nips:text_convnet} or even outperform ConvNets for some
data sets.

In summary, we are not aware of any work that uses VGG-like or ResNet-like architecture to go deeper than than six convolutional
layers \cite{Zhang:2015_nips:text_convnet} for sentence classification.  Deeper networks were not tried or they
were reported to not improve performance.  This is in sharp contrast to the
current trend in computer vision where significant improvements have been
reported using much deeper networks\cite{krizhevsky2012imagenet}, namely 19 layers~\cite{msr:2016:iclr:vgg},
or even up to 152 layers~\cite{He:2015:resnet}.
In the remainder of this paper, we describe our very deep convolutional
architecture and report results on the same corpora than
\cite{Zhang:2015_nips:text_convnet}. We were able to show that performance
improves with increased depth, using up to 29 convolutional layers.


\section{VDCNN Architecture}
\label{SectArch}

\begin{figure}
  \centering
  \includegraphics[width=0.8\linewidth]{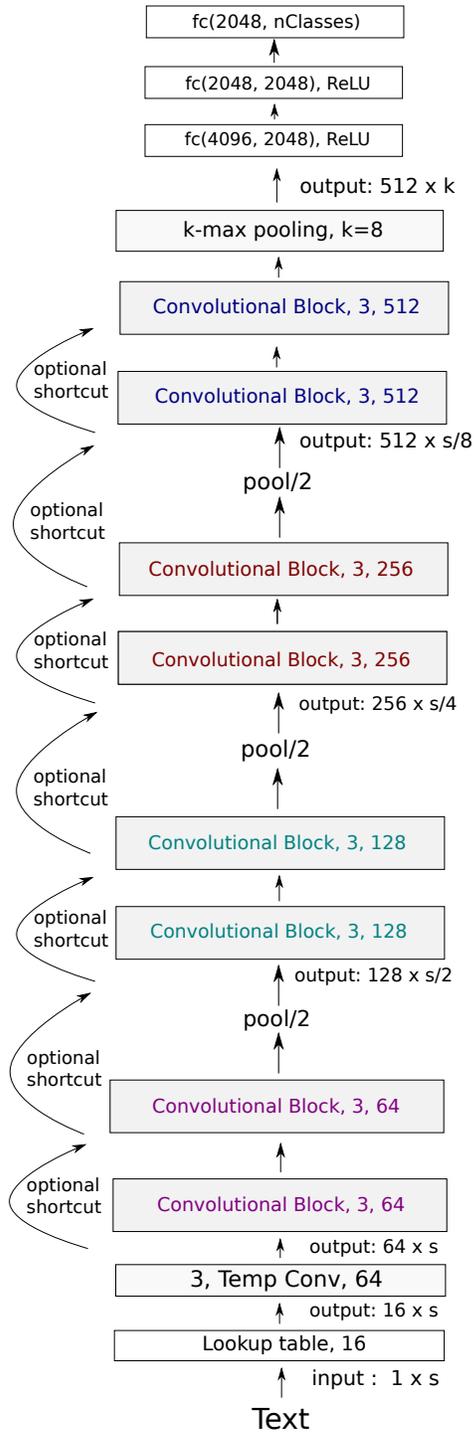}
   \caption{VDCNN architecture.}
     \label{FigArchi}

\end{figure}

The overall architecture of our network is shown in Figure~\ref{FigArchi}.  Our
model begins with a look-up table that generates a 2D tensor of size $(f_0, s)$ that contain the embeddings of the $s$ characters. $s$ is fixed to 1024, and $f_0$ can be seen as the "RGB" dimension of the input text.

We first apply one layer of 64 convolutions of size $3$, followed by
a stack of temporal ``convolutional blocks''. Inspired by the philosophy of VGG
and ResNets we apply these two design rules: (i) for the same output temporal
resolution, the layers have the same number of feature maps, (ii) when the
temporal resolution is halved, the number of feature maps is doubled. This helps reduce the memory footprint of the network.  The networks contains 3 pooling operations (halving the temporal
resolution each time by 2), resulting in 3 levels of 128, 256 and 512 feature
maps (see Figure~\ref{FigArchi}).
The output of these convolutional blocks is a tensor of size $512\times s_d$,
where $s_d=\frac{s}{2^{p}}$ with $p=3$ the number of down-sampling
operations. At this level of the convolutional network, the resulting tensor
can be seen as a high-level representation of the input text. Since we deal
with padded input text of fixed size, $s_d$ is constant. However, in the case
of variable size input, the convolutional encoder provides a representation of
the input text that depends on its initial length $s$. Representations of a
text as a set of vectors of variable size can be valuable namely for neural
machine translation, in particular when combined with an attention model.
In Figure~\ref{FigArchi},
temporal convolutions with kernel size 3 and X feature maps are denoted
"\texttt{3, Temp Conv, X}", fully connected layers which are linear projections
(matrix of size $I\times O$) are denoted "\texttt{fc(I, O)}" and "\texttt{3-max
pooling, stride 2}" means temporal max-pooling with kernel size 3 and stride 2.

Most of the previous applications of ConvNets to NLP use an architecture which is
rather shallow (up to 6 convolutional layers) and combines convolutions of
different sizes, \eg spanning 3, 5 and 7 tokens. This was motivated by the fact
that convolutions extract $n$-gram features over tokens and that different
$n$-gram lengths are needed to model short- and long-span relations.  In this
work, we propose to create instead an architecture which uses many layers of small
convolutions (size 3). Stacking 4 layers of such convolutions
results in a span of 9 tokens, but the network can learn by itself how to best
combine these different \textit{``3-gram features''} in a deep hierarchical
manner.
Our architecture can be in fact seen as a temporal adaptation of the VGG
network~\cite{msr:2016:iclr:vgg}.  We have also investigated the same kind of
\textit{``ResNet shortcut''} connections as in \cite{He:2015:resnet},
namely identity and $1\times 1$ convolutions (see Figure~\ref{FigArchi}).

For the classification tasks in this work, the temporal resolution
of the output of the convolution blocks is first down-sampled to a fixed
dimension using $k$-max pooling. By these means, the network extracts the $k$
most important features, independently of the position they appear in the
sentence.  The $512\times k$ resulting features are transformed into a single
vector which is the input to a three layer fully connected classifier with ReLU
hidden units and softmax outputs. The number of output neurons depends on the
classification task, the number of hidden units is set to 2048, and $k$ to 8 in
all experiments. We do not use drop-out with the fully connected layers, but only temporal batch
normalization after convolutional layers to regularize our network.

\subsection*{Convolutional Block}

\label{subsectConvBlock}

\begin{figure}
  \centering
  \includegraphics[width=0.5\linewidth]{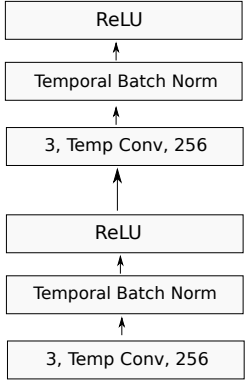}
    \caption{Convolutional block.}
      \label{FigBlock}

  \vspace{-5mm}
\end{figure}
Each convolutional block (see Figure~\ref{FigBlock}) is a sequence of two
convolutional layers, each one followed by a temporal
BatchNorm~\cite{Ioffs:2015:icml_batchnorm} layer and an ReLU activation. The
kernel size of all the temporal convolutions is 3, with padding such that
the temporal resolution is preserved (or halved in the case of the convolutional
pooling with stride 2, see below). Steadily increasing the depth of the network
by adding more convolutional layers is feasible thanks to the limited number of
parameters of very small convolutional filters in all layers. Different depths of the overall architecture
are obtained by varying the number of convolutional blocks in between the
pooling layers (see table~\ref{tabledepth}). Temporal batch
normalization applies the same kind of regularization as batch normalization
except that the activations in a mini-batch are jointly normalized over
temporal (instead of spatial) locations. So, for a mini-batch of size $m$ and
feature maps of temporal size $s$, the sum and the standard deviations related
to the BatchNorm algorithm are taken over $\vert \mathcal{B} \vert = m\cdot s$
terms.

We explore three types of down-sampling between blocks $K_i$ and $K_{i+1}$ (Figure~\ref{FigArchi}) : 
\begin{enumerate}
\item[(i)] The first
convolutional layer of $K_{i+1}$ has stride 2 (ResNet-like).
\item[(ii)] $K_i$ is followed by a $k$-max pooling layer where $k$ is such that
the resolution is halved \cite{Kalchbrenner:2014:convNNMT}.
\item[(iii)] $K_i$ is followed by max-pooling
with kernel size 3 and stride 2 (VGG-like).
\end{enumerate}
All these types of pooling reduce the temporal
resolution by a factor 2. At the final convolutional layer, the resolution is
thus $s_d$.

\begin{table}[!ht]
  \centering
  \begin{tabular}{|lcccc|}
    \hline
    Depth: &  9 & 17 & 29 & 49\\
    \hline
    conv block 512 & 2 & 4 & 4 & 6 \\
    conv block 256 & 2 & 4 & 4 & 10 \\
    conv block 128 & 2 & 4 & 10 & 16 \\
    conv block 64 & 2 & 4 & 10 & 16 \\
    First conv. layer & 1 & 1 & 1 & 1 \\
    \hline
    \#params [in M] & 2.2 & 4.3 & 4.6 & 7.8 \\
    \hline
  \end{tabular}
    \caption{Number of conv. layers per depth.}
      \label{tabledepth}
  \vspace{-0.5cm}
\end{table}
In this work, we have explored four depths for our networks: 9, 17, 29 and 49, which
we define as being the number of convolutional layers. The depth of a network
is obtained by summing the number of blocks with 64, 128, 256 and 512 filters,
with each block containing two convolutional layers. In Figure~\ref{FigArchi}, the
network has 2 blocks of each type, resulting in a depth of
$2\times(2+2+2+2)=16$. Adding the very first convolutional layer, this sums to
a depth of 17 convolutional layers. The depth can thus be increased or
decreased by adding or removing convolutional blocks with a certain number
of filters. The best configurations we observed for depths 9, 17, 29 and 49 are
described in Table~\ref{tabledepth}.
We also give the number of parameters of all convolutional layers.

\begin{table*}[!ht]
  \centering
  \begin{tabular}{|lrrrl|}
    \hline
    Data set     & \#Train & \#Test & \#Classes & Classification Task \\
    \hline
    AG's news & 120k & 7.6k & 4 & English news categorization \\
    Sogou news  & 450k & 60k & 5  & Chinese news categorization \\
    DBPedia  & 560k & 70k & 14 & Ontology classification \\
    Yelp Review Polarity  & 560k & 38k & 2  &  Sentiment analysis \\
    Yelp Review Full & 650k & 50k & 5 & Sentiment analysis \\
    Yahoo! Answers & 1~400k &  60k & 10 & Topic classification \\
    Amazon Review Full &  3~000k & 650k & 5 & Sentiment analysis \\
    Amazon Review Polarity  & 3~600k & 400k &  2 & Sentiment analysis \\
    \hline
  \end{tabular}
    \caption{Large-scale text classification data sets used in our experiments.
	See \cite{Zhang:2015_nips:text_convnet} for a detailed description.}
	  \label{figdatasets}
\end{table*}

\section{Experimental evaluation}
\label{SectExp}

\subsection{Tasks and data}
\label{SectData}


In the computer vision community, the availability of large data sets for
object detection and image classification has fueled the development of new
architectures. In particular, this made it possible to compare many different
architectures and to show the benefit of very deep convolutional networks.
We present our results on eight freely available large-scale data sets
introduced by \cite{Zhang:2015_nips:text_convnet} which cover several
classification tasks such as sentiment analysis, topic classification or news
categorization (see Table~\ref{figdatasets}). The number of training examples
varies from 120k up to 3.6M, and the number of classes is comprised between 2
and 14.  This is considerably lower than in computer vision (\eg 1~000 classes
for ImageNet). This has the consequence that each example induces less
gradient information which may make it harder to train large architectures.  It should be
also noted that some of the tasks are very ambiguous, in particular sentiment
analysis for which it is difficult to clearly associate fine grained labels.
There are equal numbers of examples in each class for both training and test
sets.  The reader is referred to \cite{Zhang:2015_nips:text_convnet} for more
details on the construction of the data sets.
Table~\ref{FigRefResults} summarizes the best published results on these corpora
we are aware of.
We do not use ``Thesaurus data augmentation'' or
any other preprocessing, except lower-casing. Nevertheless, we still
outperform the best convolutional neural networks of
\cite{Zhang:2015_nips:text_convnet} for all data sets.  The main goal of our
work is to show that it is possible and beneficial to train very deep
convolutional networks as text encoders. Data augmentation may improve our
results even further.  We will investigate this in future research.

\subsection{Common model settings}

The following settings have been used in all our experiments. They were found
to be best in initial experiments.  Following
\cite{Zhang:2015_nips:text_convnet}, all processing is done at the character
level which is the atomic representation of a sentence, same as pixels for
images.  The dictionary consists of the following characters
"\texttt{abcdefghijklmnopqrstuvwxyz0123456}\\\texttt{789-,;.!?:'"/|\_\@\#\$\%\^{
}\&*\~{}`+=<>()[]\{\}}" plus a special padding, space and unknown token which add up
to a total of 69 tokens.  The input text is padded to a fixed size of 1014, larger text are truncated.
The character embedding is of size 16.
Training is performed with SGD, using a mini-batch of size 128, an initial
learning rate of 0.01 and momentum of 0.9.  We follow the same training procedure as in \cite{Zhang:2015_nips:text_convnet}.  We initialize our convolutional
layers following \cite{He:2015:iccv:imgnet_init}. One epoch took from 24 minutes to 2h45 for depth 9, and from 50 minutes to 7h (on the largest datasets) for depth 29. It took between 10 to 15 epoches to converge.  The implementation is done
using Torch 7.  All experiments are performed on a single NVidia K40 GPU.  Unlike previous research on the use of ConvNets for text
processing, we use temporal batch norm without dropout.

\begin{table*}[!ht]
  \centering
  \begin{tabular}{|lcccccccc|}
    \hline
    Corpus: &  AG & Sogou & DBP. & Yelp P. & Yelp F. & Yah. A. & Amz. F. & Amz. P. \\
    \hline
    Method & n-TFIDF & n-TFIDF & n-TFIDF & ngrams & Conv & Conv+RNN & Conv & Conv \\
    Author & [Zhang] & [Zhang] & [Zhang] & [Zhang] & [Zhang] & [Xiao] & [Zhang] & [Zhang] \\

    Error & 7.64 & 2.81 & 1.31 & 4.36 & 37.95$^*$ & 28.26 & 40.43$^*$ & 4.93$^*$ \\
    $[$Yang$]$ & - & - & - & - & - & 24.2 & 36.4 & - \\
    \hline
    
  \end{tabular}
  \vspace{-3mm}
  \caption{Best published results from previous work.
     \newcite{Zhang:2015_nips:text_convnet} best results use a Thesaurus data augmentation technique
     (marked with an $^*$).
     \newcite{yang2016hierarchical}'s hierarchical methods is particularly adapted to datasets whose samples contain multiple sentences.}
       \label{FigRefResults}
\end{table*}

\subsection{Experimental results}

\begin{table*}[!ht]
  \centering
  \begin{tabular}{|cl*{8}{@{\hspace{1.6mm}}c@{\hspace{1.6mm}}}|}
    \hline
    Depth & Pooling & AG & Sogou & DBP. & Yelp P. & Yelp F. & Yah. A. & Amz. F. & Amz. P. \\
    \hline
    9 & Convolution & 10.17 & 4.22 & 1.64 & 5.01 & 37.63 & 28.10 & 38.52 & 4.94 \\
    9 & KMaxPooling &  9.83 & 3.58 & 1.56 & 5.27 & 38.04 & 28.24 & 39.19 & 5.69 \\
    9 & MaxPooling  &  9.17 & 3.70 & 1.35 & 4.88 & 36.73 & 27.60 & 37.95 & 4.70  \\
    \hline
    17 & Convolution & 9.29 & 3.94 & 1.42 & 4.96 & 36.10 & 27.35 & 37.50 & 4.53 \\
    17 & KMaxPooling & 9.39 & 3.51 & 1.61 & 5.05 & 37.41 & 28.25 & 38.81 & 5.43 \\
    17 & MaxPooling  & 8.88 & 3.54 & 1.40 & 4.50 & 36.07 & 27.51 & 37.39 & 4.41  \\
    \hline
    29 & Convolution & 9.36 & 3.61 & 1.36 & 4.35 & \textbf{35.28} & 27.17 & 37.58 & \textbf{4.28} \\
    29 & KMaxPooling & \textbf{8.67} & \textbf{3.18} & 1.41 & 4.63 & 37.00 & 27.16 & 38.39 & 4.94 \\
    29 & MaxPooling  & 8.73 & 3.36 & \textbf{1.29} & \textbf{4.28} & 35.74 & \textbf{26.57} & \textbf{37.00} & 4.31  \\
    \hline
  \end{tabular}
    \caption{Testing error of our models on the 8 data sets. No data preprocessing or augmentation is used.}
   \label{figresults}
    \vspace{-5mm}

\end{table*}

In this section, we evaluate several configurations of our model, namely three
different depths and three different pooling types (see Section~\ref{subsectConvBlock}).
Our main contribution is a
thorough evaluation of networks of increasing depth using an architecture with
small temporal convolution filters with different types of pooling, which shows
that a significant improvement on the state-of-the-art configurations can be
achieved on text classification tasks by pushing the depth to 29 convolutional
layers.

\paragraph{Our deep architecture works well on big data sets in particular, even for small depths.}
Table \ref{figresults} shows the test errors for depths 9, 17 and 29 and for
each type of pooling : convolution with stride 2, $k$-max pooling and temporal
max-pooling. For the smallest depth we use (9 convolutional layers), we see that our model already
performs better than Zhang's convolutional baselines (which includes 6 convolutional layers
and has a different architecture) on the biggest data sets :
Yelp Full, Yahoo Answers and Amazon Full and Polarity. The most important decrease in
classification error can be observed on the largest data set Amazon Full which
has more than 3 Million training samples. We also observe that for a small
depth, temporal max-pooling works best on all data sets. 

\paragraph{Depth improves performance.}
As we increase the network depth to 17 and 29, the test errors decrease on all data sets,
for all types of pooling (with 2 exceptions for 48 comparisons). Going from depth 9 to 17 and 29 for Amazon Full
reduces the error rate by 1\% absolute. Since the test is composed of
650K samples, 6.5K more test samples have been classified correctly. These improvements,
especially on large data sets, are significant and show that increasing the
depth is useful for text processing. Overall, compared to previous
state-of-the-art, our best architecture with depth 29 and max-pooling has a
test error of 37.0 compared to 40.43\%. This represents a gain
of 3.43\% absolute accuracy.
The significant improvements which we obtain on all
data sets compared to Zhang's convolutional models do not include any data
augmentation technique.

\paragraph{Max-pooling performs better than other pooling types.}
In terms of pooling, we can also see that max-pooling performs best overall,
very close to convolutions with stride 2, but both are significantly superior to
$k$-max pooling.

Both pooling mechanisms perform a max operation which is local and limited to
three consecutive tokens, while $k$-max polling considers the whole sentence at
once.  According to our experiments, it seems to hurt performance to perform
this type of max operation at intermediate layers (with the exception of the
smallest data sets).


\paragraph{Our models outperform state-of-the-art ConvNets.}
We obtain state-of-the-art results for all data sets, except AG's news and Sogou
news which are the smallest ones. However, with our very deep
architecture, we get closer to the state-of-the-art which are
ngrams TF-IDF for these data sets and significantly surpass convolutional models
presented in \cite{Zhang:2015_nips:text_convnet}.
As observed in previous work, differences in accuracy between shallow
(TF-IDF) and deep (convolutional) models are more significant on large data
sets, but we still perform well on small data sets while getting closer to the
non convolutional state-of-the-art results on small data sets. The very deep models even
perform as well as ngrams and ngrams-TF-IDF respectively on the sentiment
analysis task of Yelp Review Polarity and the ontology classification task of
the DBPedia data set. Results of Yang et al. (only on Yahoo Answers and Amazon Full) outperform our model on the Yahoo Answers dataset, which is probably linked to the fact that their model is task-specific to datasets whose samples that contain multiple sentences like (question, answer). They use a hierarchical attention mechanism that apply very well to documents (with multiple sentences).

\paragraph{Going even deeper degrades accuracy. Shortcut connections help reduce the degradation.}
As described in \cite{He:2015:resnet}, the gain in accuracy due to the
the increase of the depth is limited when using standard ConvNets. When the
depth increases too much, the accuracy of the model gets saturated and starts
degrading rapidly. This \textit{degradation} problem was attributed to the fact
that very deep models are harder to optimize. The gradients which are
backpropagated through the very deep networks vanish and SGD with momentum is
not able to converge to a correct minimum of the loss function. To overcome this
degradation of the model, the \textit{ResNet model} introduced shortcut
connections between convolutional blocks that allow the gradients to flow more
easily in the network~\cite{He:2015:resnet}. 

We evaluate the impact of shortcut connections by increasing the number of
convolutions to 49 layers. We present an adaptation of the ResNet model to the
case of temporal convolutions for text (see Figure~\ref{FigArchi}).
Table~\ref{figresnet} shows the evolution of the test errors on the Yelp
Review Full data set with or without shortcut connections. When looking at the column ``without shortcut'', we observe the
same degradation problem as in the original ResNet article: when going from 29
to 49 layers, the test error rate
increases from 35.28 to 37.41 (while the training error goes up from 29.57 to 35.54).  When using shortcut connections,
we observe improved results when the network has 49 layers: both the training
and test errors go down and the network is less prone to underfitting than it
was without shortcut connections.

While shortcut connections give better results when the network is very deep
(49 layers), we were not able to reach state-of-the-art results with them. We
plan to further explore adaptations of residual networks to temporal
convolutions as we think this a milestone for going deeper in NLP. Residual units \cite{He:2015:resnet} better adapted to the text processing task may help for training even deeper models for text processing, and is left for future research.

\begin{table}[!ht]
  \centering
  \begin{tabular}{|c|cc|}
    \hline
     depth& without shortcut & with shortcut \\
         \hline
    9 & 37.63 & 40.27 \\
    17 & 36.10 & 39.18 \\
    29 & 35.28 & 36.01  \\
    49 & 37.41 & 36.15 \\
    \hline
  \end{tabular}
    \caption{Test error on the Yelp Full data set
           for all depths, with or without residual connections.}
     \label{figresnet}
    \vspace{-3mm}

\end{table}

\paragraph{Exploring these models on text classification tasks with more classes sounds promising.}
Note that one of the most important difference between the classification tasks
discussed in this work and ImageNet is that the latter deals with 1000
classes and thus much more information is back-propagated to the network through
the gradients.  Exploring the impact
of the depth of temporal convolutional models on categorization tasks with
hundreds or thousands of classes would be an interesting challenge and is left
for future research.

\section{Conclusion}
We have presented a new architecture for NLP which follows two design principles:
1) operate at the lowest atomic representation of text, i.e. characters,
and 2) use a deep stack of local operations, i.e. convolutions and max-pooling of size 3,
   to learn a high-level hierarchical representation of a sentence.
This architecture has been evaluated on eight freely available large-scale data
sets and we were able to show that increasing the depth up to 29 convolutional
layers steadily improves performance.  Our models are much deeper than
previously published convolutional neural networks and they outperform those
approaches on all data sets.  To the best of our knowledge, this is the first
time that the \textit{``benefit of depths''} was shown for convolutional neural
networks in NLP.

Eventhough text follows human-defined rules and images can be seen as raw signals of our environment, images and small
texts have similar properties. Texts are also compositional for many languages.
Characters combine to form n-grams, stems, words, phrase, sentences etc. These similar
properties make the comparison between computer vision and natural language
processing very profitable and we believe future research should invest into
making text processing models deeper. Our work is a first attempt towards this
goal.

In this paper, we focus on the use of very deep convolutional neural networks
for sentence classification tasks. Applying similar ideas to other sequence
processing tasks, in particular neural machine translation is left for future research. It needs to be investigated
whether these also benefit from having deeper convolutional encoders.
\newpage

%

\bibliography{eacl2017}
\bibliographystyle{eacl2017}

\end{document}